\documentclass{article}

\usepackage{PRIMEarxiv}

\usepackage[utf8]{inputenc} 
\usepackage[T1]{fontenc}    
\usepackage{url}            
\usepackage{booktabs}       
\usepackage{amsfonts}       
\usepackage{nicefrac}       
\usepackage{microtype}      
\usepackage{lipsum}
\usepackage{fancyhdr}       
\usepackage{graphicx}       
\graphicspath{{media/}}     

\def\etal{\emph{et al.}~}

\usepackage{float}
\usepackage{amssymb}
\usepackage{lipsum}
\usepackage{xurl}
\usepackage[colorlinks = true,
            linkcolor = blue,
            urlcolor  = magenta,
            citecolor = blue,
            anchorcolor = blue]{hyperref}
\title{Neural Style Transfer for Computer Games
}

\author{
  Eleftherios Ioannou, Steve Maddock \\
  Department of Computer Science \\
  University of Sheffield \\
  Sheffield, UK \\
  \texttt{\{eioannou1, s.maddock\}@sheffield.ac.uk} \\
}

\begin{document}
\maketitle

\begin{abstract}
Neural Style Transfer (NST) research has been applied to images, videos, 3D meshes and radiance fields, but its application to 3D computer games remains relatively unexplored. Whilst image and video NST systems can be used as a post-processing effect for a computer game, this results in undesired artefacts and diminished post-processing effects. Here, we present an approach for injecting depth-aware NST as part of the 3D rendering pipeline. Qualitative and quantitative experiments are used to validate our in-game stylisation framework. We demonstrate temporally consistent results of artistically stylised game scenes, outperforming state-of-the-art image and video NST methods. 
\end{abstract}

\begin{center}
    \url{https://ioannoue.github.io/nst-for-computer-games.html}
\end{center}


\begin{figure*}[htb]
    \begin{center}
        \includegraphics[width=\linewidth]{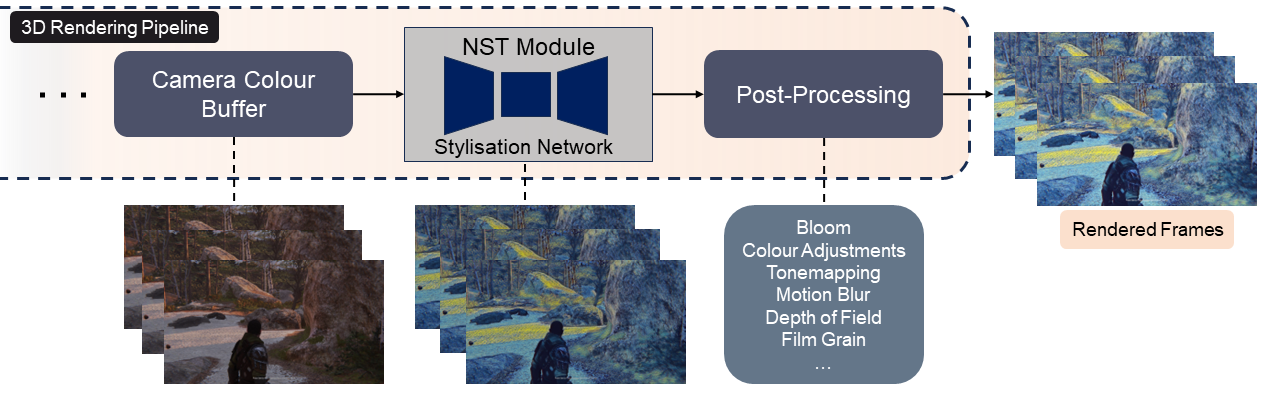}
    \end{center}
   \caption{Our proposed framework injects NST as part of the 3D rendering pipeline.} 
\label{fig:teaser}
\end{figure*}

\section{Introduction}

Neural Style Transfer (NST) refers to the process of changing the appearance of an input image based on a reference style image, whilst preserving the input image's underlying content.
For example, a photograph of a landscape can be made to take on the style of a Van Gogh painting. More recently, NST has been extended to work for three-dimensional data, such as 3D meshes, point clouds, and radiance fields. Given the inherently creative and artistic nature of NST, another domain where its application holds immense potential is within the realm of 3D computer games. By integrating style transfer techniques into computer games, developers could dynamically alter the visual aesthetics of a game in real-time, and players could be given the ability to choose from an array of artistic styles, influencing the appearance of the game's world and characters according to their preferences.

However, there is limited work applying NST to 3D computer games.
Whilst NST approaches have not been specifically tailored for 3D computer games, image and video NST methods can be applied at the end of the 3D computer graphics pipeline, as a post-processing effect \cite{unity_barracuda}. This essentially treats the data as a sequence of images. Here, temporal consistency across consecutive frames is the prominent challenge. Some video NST approaches utilise optical flow information and introduce a temporal consistency loss to achieve temporal stability \cite{huang2017real,ruder2018artistic,gao2018reconet,chen2016fast,gao2020fast}, whilst other approaches rely on improving the stability of the transferred content and style features \cite{liu2021adaattn,deng2021arbitrary,lu2022universal}. Nevertheless, employing such models at the post-process stage of the 3D computer graphics pipeline, results in undesired flickering effects and inconsistent stylisations.

Previous work has demonstrated that the utilisation of G-buffer data can lead to improved quality of generated stylised game scenes \cite{richter2022enhancing,Mittermueller2022estgan}.
Our work takes advantage of the intermediate data that is generated by a 3D computer graphics rendering pipeline, and proposes an approach for integrating an NST model at an earlier stage of the rendering process (Figure~\ref{fig:teaser}), resulting in improved, more stable artistic stylisations of game worlds. Our method retrieves data from the camera colour buffer, generates consistent stylised game frames, and writes back to the colour buffer, before post-processing. We believe this is the first work to stylise in this way. The primary contributions of our work can be summarised as follows: 1) We train a fast Stylisation network on both a real-world and a synthetic image dataset, capable of producing fast high-quality artistic stylisations; 2) We present an approach that integrates a trained stylisation network at an early stage of the rendering pipeline, avoiding the visual artefacts and inconsistencies that occur when employing stylisation as a post-effect; 3) We evaluate the results of our system qualitatively and quantitatively, demonstrating how the games community can benefit from the NST field.

\section{Related Work}
\label{sec:related_work}

\label{sec:related-work}

\subsection{Image \& Video NST}

Gatys~\etal \cite{gatys2016image} proposed a model that minimises a content loss and a style loss, based on features extracted from pre-trained CNNs on object classification. Since this seminal work, multiple NST approaches have emerged, proposing end-to-end systems trained on singular styles that manage to improve upon the efficiency and time required to generate one stylisation.
These are capable of synthesising a stylised output with a single forward pass through the network \cite{johnson2016perceptual, ulyanov2016texture, ulyanov2017improved,li2016precomputed}. While some models trained to capture the style of a particular artist or a specific art genre \cite{sanakoyeu2018style, kotovenko2019content}, and more efficient multiple-style-per-model approaches were also developed \cite{dumoulin2017learned, chen2017stylebank, zhang2018multi}, recently, the research has shifted to developing arbitrary style transfer systems. The method of Huang and Belongie \cite{huang2017arbitrary} suggested the use of an Adaptive Instance Normalisation layer (AdaIN) that allows transferring the channel-wise mean and variance feature statistics between the content and style feature activations, thus achieving arbitrary style transfer. Other arbitrary-style-per-model methods were also developed that improve upon the performance \cite{ghiasi2017exploring,li2017universal,xu2018learning,huo2021manifold,park2019arbitrary,svoboda2020two,an2020ultrafast} or offer solutions tailored to particular challenges \cite{hu2020aesthetic, liu2021learning}. Meta networks were also employed \cite{shen2018neural}, as well as systems making use of the recently developed transformer architecture \cite{liu2021adaattn,deng2022stytr2,luo2022consistent}.


To alleviate the issue of temporal inconsistency across subsequent frames when video is considered, Ruder~\etal \cite{ruder2016artistic,ruder2018artistic} employed a temporal constraint based on optical flow information. Typically, the optical flow map (calculated between two frames of the original video) is used to warp the previous stylised frame to give an estimation of the next frame. This gives a temporal loss function that can be minimised during training. 
Other work has subsequently improved the computation speed \cite{huang2017real,gao2018reconet} or demonstrated structure and depth-preserving qualities \cite{liu2021structure,ioannou2023depth}. Gao~\etal \cite{gao2020fast} developed a fast model that incorporates multiple styles, while arbitrary video style transfer models have also been proposed \cite{wang2020consistent,deng2021arbitrary,lu2022universal,wu2022ccpl}, some of which are extensions from image NST approaches with additional temporal considerations \cite{li2019learning,liu2021adaattn,liu2021structure}.

\subsection{NST in 3D Computer Games}

Although methods exist for stylising three-dimensional data and can offer 3D artists diverse options for generating or improving a game's assets, no substantial efforts have been noted for real-time in-game artistic stylisation. The image and video NST models \cite{liu2021adaattn,deng2021arbitrary,lu2022universal,gao2020fast} can potentially be integrated at the end of a computer game's rendering pipeline, intercepting each rendered frame and producing a stylised version of it.  An example of this has been exhibited by Unity's implementation \cite{deliot_guinier_vanhoey_2020} which is based on the method of Ghiasi~\etal \cite{ghiasi2017exploring} that produces a stylised image from an input image in a single forward pass from the neural network. Multi-style in-game style transfer is achieved allowing the viewer to change the stylisation of the scene in real-time. Nonetheless, the implementation does not consider any G-buffer or 3D information. Instead, it intercepts the final rendered 2D image (using an off-screen buffer), which means it can be applied as a final ‘filter’ for any game. This also results in unstable stylisations and causes the intended post-process effects being diminished. 

The recent approach by Richter~\etal \cite{richter2022enhancing} to enhancing the photorealism of computer-generated imagery might be the first to take into account information from intermediate buffers (G-buffers) that becomes available through a game engine’s rendering process. This method, although explicitly focused on photorealistic enhancement, can be significantly impactful to style transfer algorithms that consider the stylisation of game environments. Their technique also works at the end of the rendering pipeline -- the image enhancement network outputs an enhanced image given an input rendered image. However, the image enhancement network is fed with information about the geometry, materials, and lighting, extracted from intermediate rendering buffers during training.  

Similarly, the image-to-image translation method proposed by Mittermueller~\etal \cite{Mittermueller2022estgan} trains a network to learn the mapping between low-poly game scenes to a synthetic dataset compiled using the Red Dead Redemption 2 (RDR2) game. The mapping takes into account intermediate data such as depth, normals, and albedo generated by conventional game rendering pipelines, for improved image domain transfer.  Although the developed \textit{EST-GAN} validates the effectiveness of G-buffer data for the generation of stylistic game scenes, it does not utilise this information in real-time and does not demonstrate any impact on the stability of sequential stylised game frames.

Integrating NST at the end of the 3D rendering pipeline is the only approach that has been suggested for the synthesis of real-time photorealistic \cite{richter2022enhancing} or artistic \cite{deliot_guinier_vanhoey_2020} game worlds.
Nevertheless, the amount of post-processing that is executed, and the unpredictable camera movement and scene shifts, do not allow for coherent and robust post-process stylisations. Here we propose an approach for producing stable and aesthetically pleasing visual effects in computer games by integrating a style transfer model before the post-process stage of the 3D computer game rendering pipeline.


\section{Injecting NST into the 3D rendering pipeline}
\label{sec:method}

\subsection{Style Transfer Network}

\begin{figure*}[htb]
    \begin{center}
        \includegraphics[width=\linewidth]{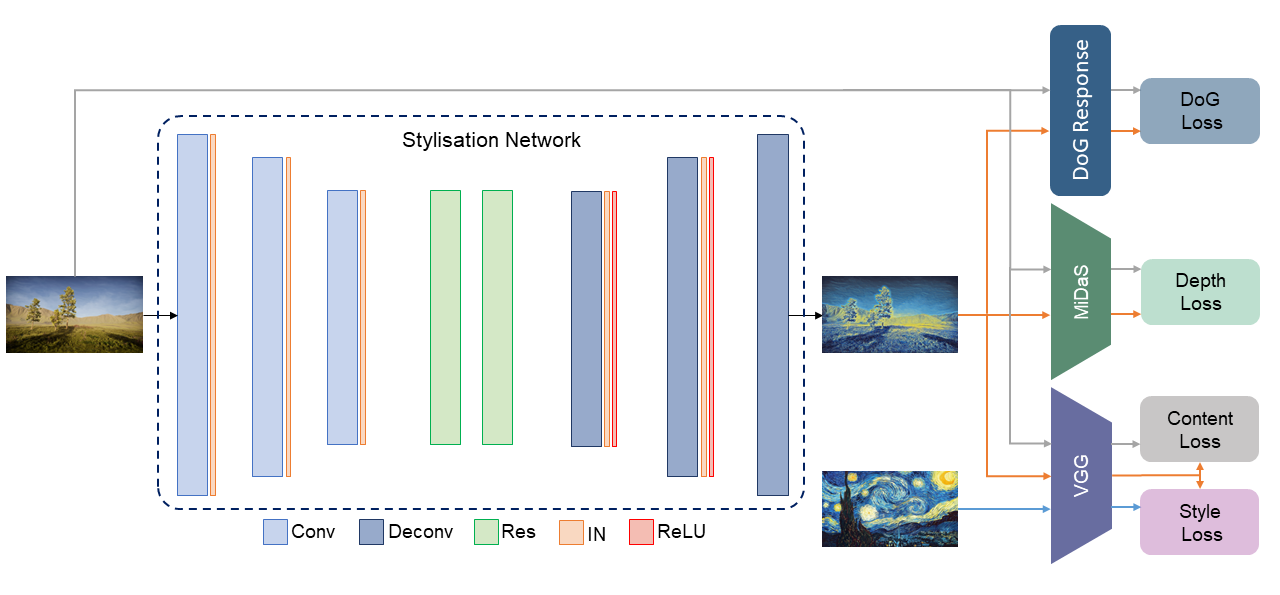}
    \end{center}
   \caption{The Stylisation Network consists of three convolutional layers (Conv), two residual layers (Res) and three deconvolutional layers (Deconv). Instance normalisation layers (IN) and the ReLU activation function are included at the first two deconvolution layers.} 
\label{fig:nst-network}
\end{figure*}

The network architecture is shown in Figure~\ref{fig:nst-network}. Similarly to state-of-the-art methods \cite{johnson2016perceptual,liu2017depth,ioannou2022depth}, we utilise a Transformation network $f_{W}$, that intercepts an input image $x$ and transforms it into an output image $\hat{y}$ via the mapping $\hat{y} = f_{W}(x)$. To improve upon the efficiency and inference time required to generate a stylised frame given an input image, we reduce the number of residual layers and remove the ReLU activation function from the first three convolutional layers. The final configuration of our network consists of three convolutional layers followed by instance normalisation, two residual layers (composed of convolutions, instance normalisation, and ReLU), and three deconvolutional layers that upsample the input and then perform convolution. The first two deconvolutional layers are followed by instance normalisation and ReLU activation.

\subsubsection{Content \& Style Losses}

We use the perceptual loss functions introduced in the work of Johnson~\etal \cite{johnson2016perceptual} and employ a pre-trained image recognition network (\textit{VGG-16}~\cite{simonyan2015deep}) to produce feature representations of the original and transformed images. Content loss is defined as the Euclidean distance between the feature representations of the input image and the corresponding transformed image, as extracted from the $relu2\_2$ layer:

\begin{equation} \label{eq:contentloss}
    l^{\phi_{0}}_{content} (\hat{y}, x) = \frac{1}{C_j H_j W_j}  \| \phi^{j}_{0}(\hat{y}) - \phi^{j}_{0}(x) \|^{2}_{2}
\end{equation}
where $\phi_{0}$ is the image classification network, $\phi^{j}_{0}$ represents the activations of the $j^{th}$ layer of $\phi_{0}$, and $H \times W \times C$ is the shape of the processed image. 

The style is represented by features extracted from multiple layers of \textit{VGG-16} (\textit{J = \{relu1\_2, relu2\_2, relu3\_3, relu4\_3\}}). The Gram matrix $G$ is then computed to give feature correlations that can be utilised to define the style loss function. This is then defined as the squared Frobenius norm of the difference between the calculated Gram-based style representations:

\begin{equation} \label{eq:style-loss}
    \mathcal{L}^{\phi_{0}, j}_{style} (\hat{y}, y)  = \|  G^{\phi_{0}}_{j} (\hat{y}) - G^{\phi_{0}}_{j} (y) \|^{2}_{F}
\end{equation}
and it is summed up for all the layers $j$ in $J$. Here, $y$ and $\hat{y}$ refer to the original style image and the transformed image, respectively.

\subsubsection{Depth Loss}
Previous approaches that consider depth information during training \cite{liu2017depth,ioannou2022depth,ioannou2023depth} have shown improvements to the synthesised results in terms of structure retainment and depth preservation performance. As the trained stylisation network is required to be used in a game setting -- and it is highly desired to sustain the depth of the stylised game frames -- we utilise a depth reconstruction network (MiDaS) \cite{Ranftl2020} to define a depth reconstruction loss \cite{ioannou2022depth,ioannou2023depth}:

\begin{equation} \label{eq:depth-loss}
    \mathcal{L}^{MiDaS}_{depth} (\hat{y}, x)  =  \| MiDaS_{1}(\hat{y}) - MiDaS_{1}(x) \|^{2}_{2}
\end{equation}

\subsubsection{Difference of Gaussians Loss}
A particular issue that occurs in stylisation approaches is an undesired halo effect around distinct parts of an image. This effect is compounded by the significance that is placed on edges in human vision \cite{palmer1999vision,marr1980theory} meaning that edge inconsistencies stand out. We use the Difference-of-Gaussians (DoG) operator in order to improve upon the global and local structure preservation of stylised image frames, and thus attempt to alleviate the issue of the undesired halo effect. 

Inspired by the neural processing in the retina of the human eye, the DoG response is equivalent to a band-pass filter that discards most of the spatial frequencies that are present in an image. The DoG operator is derived from the concept of convolving an image with two Gaussian kernels of different standard deviations and then taking the difference between the two convolved images. This feature enhancement algorithm has been shown to produce aesthetic edge lines and has been previously utilised for image stylisation \cite{winnemoller2012xdog}. We, therefore, define a DoG loss that is based on the difference between the DoG responses (DoGR) of the original image $x$ and the corresponding stylised image $\hat{y}$: 

\begin{equation} \label{eq:dog-loss}
    \mathcal{L}_{DoG} (\hat{y}, x)  = \|  DoGR (\hat{y}) - DoGR (x) \|^{2}_{2}
\end{equation}

\subsubsection{Training Details}

The Stylisation Network is trained for 2 epochs with a batch size of 2 and a learning rate of $1 \times 10^{-3}$. The content and style weights are set to $1 \times 10^{5}$ and $1 \times 10^{10}$, respectively. The weight for the depth loss and the DoG loss is set to $1 \times 10^{3}$. The Adam optimizer~\cite{kingma2014adam} is employed with a learning rate of $1 \times 10^{-3}$. The setting of the hyperparameters is adopted from \cite{ioannou2022depth} -- this maintains the optimal content-style ratio as in the implementation of Johnson~\etal \cite{johnson2016perceptual}. To accommodate robust stylisation of game environments and synthetic scenes, both real-world images and frames from computer-generated sources are used to train the stylisation network. The MS COCO dataset \cite{lin2014microsoft} is used, mixed with frames from the MPI Sintel training set \cite{Butler2012sintel}. 
The data is shuffled and all the images are resized to $360 \times 360$ during training. In order for the trained Stylisation network to be suitable for in-game stylisation, we export the trained model to the ONNX \cite{onnx_2019} format. This is supported by Unity and the Barracuda package \cite{unity_barracuda}.

\begin{figure*}[htb]
    \begin{center}
        \includegraphics[width=\linewidth]{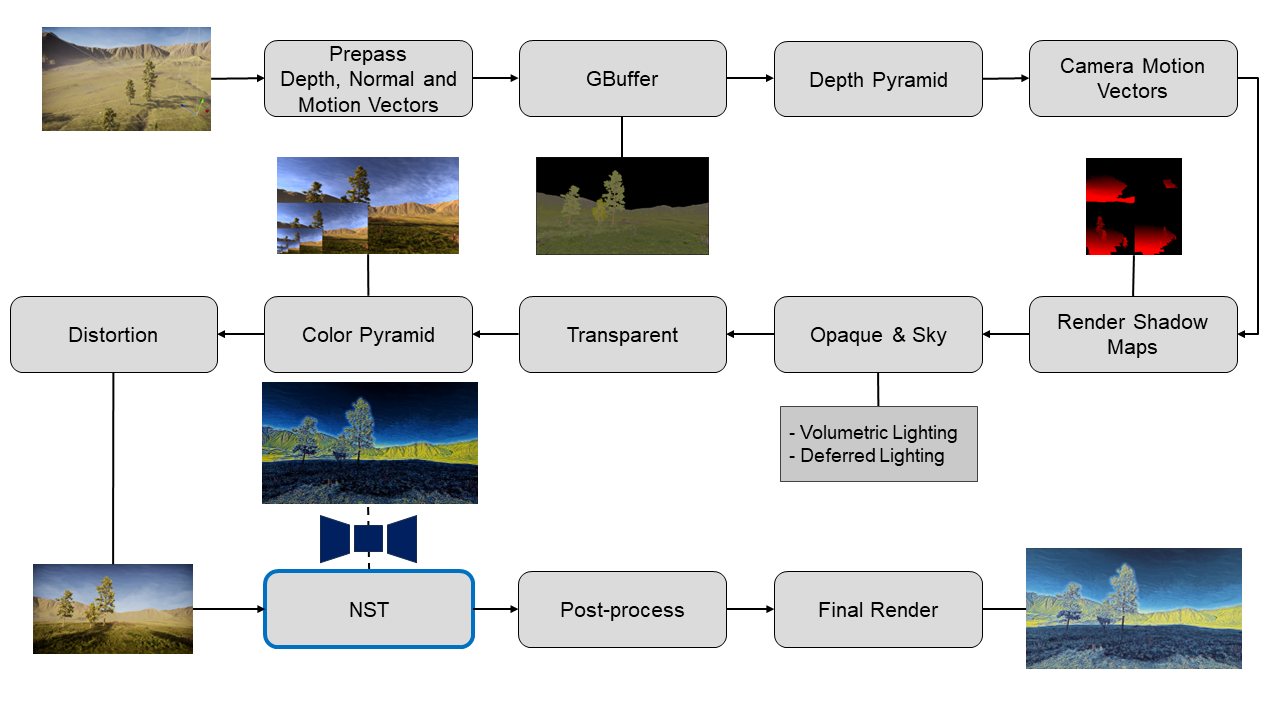}
    \end{center}
   \caption{Overview of the modified 3D rendering pipeline. The NST module is added before the Post-process stage whilst the colour buffer information is available for read and write.}
\label{fig:pipeline}
\end{figure*}

\subsection{In-game Stylisation}

To accommodate real-time in-game stylisation we use the Unity game engine and the High Definition Rendering Pipeline (HDRP) \cite{unity_hdrp}. Custom Passes can be configured within Unity's rendering pipeline and can be executed at certain points during the HDRP render loop. Six injection points for a Custom Pass are offered, with a selection of buffers being available at each. 
The injection points are: \textit{Before Rendering},  \textit{After Opaque, Depth and Normal},  \textit{Before Pre-Refraction},  \textit{Before Transparent},  \textit{Before Post-Process}, and  \textit{After Post Process}. To generate a stylised image, it is necessary to read and write to the colour buffer that is available after the  \textit{Opaque, Depth and Normal} stage. In order for the stylisation to affect the transparent objects in the scene, we opt to inject the custom pass before the Post-Process stage.

The overall modified Unity HDRP rendering pipeline is depicted in Figure~\ref{fig:pipeline}. During rendering, HDRP writes colour data from visible objects (renderers) in the scene to the colour buffer. During a custom pass, a depth pyramid and a colour pyramid are created (as shown in Figure~\ref{fig:pipeline}). The colour pyramid constitutes an iterative series of mipmaps, crafted by the HDRP, extracted from the colour buffer at a specific juncture within the rendering pipeline. The NST Module is inserted after the Distortion stage and before the Post-process stage, intercepting the colour buffer mipmap and producing an artistic stylisation for each frame (this is supported by the Barracuda package \cite{unity_barracuda} that allows for neural network inference). The synthesised texture is then passed to a custom compute shader that writes the colour to the camera colour buffer. This allows for the Post-process stage to utilise the stylised frames, before the final render. Our proposed system is capable of producing stable real-time stylised frames free from undesired artefacts and flickering effects. Embedding the NST module earlier in the rendering pipeline also allows for the post-process effects (such as depth of field, bloom, or motion blur) to effectively be visible, adding to the look and feel of the game. Such effects would be diminished if the stylisation effect was applied at the final render -- examples of this are shown in Figure~\ref{fig:results_comparison_2}.

\section{Results \& Discussion}
\label{sec:evaluation}

\begin{figure*}[htb]
    \begin{center}
        \includegraphics[width=\linewidth]{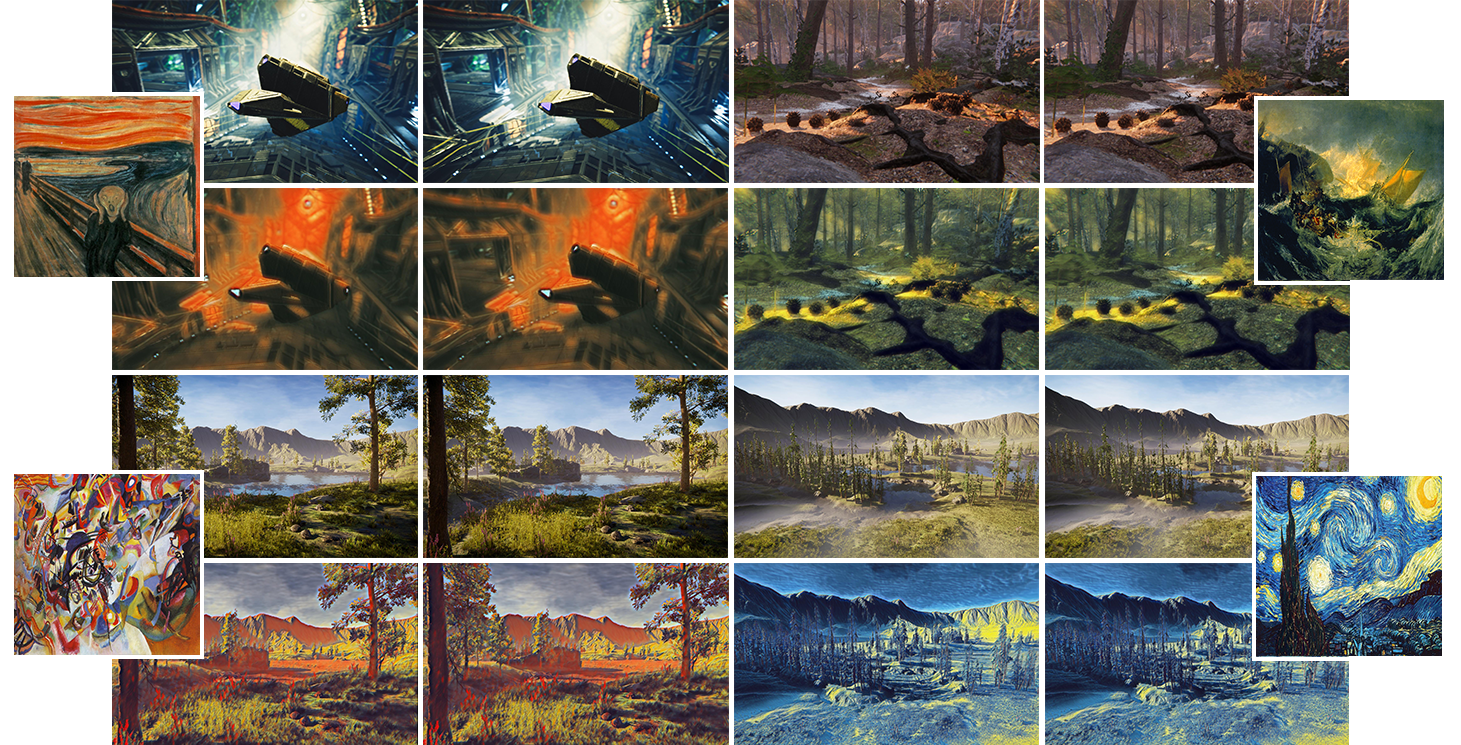}
    \end{center}
   \caption{Our approach for different style images and different game scenes. Original content images are above stylised frames. Adjacent frames show temporal stability.}
\label{fig:our_results}
\end{figure*}

Our NST system is embedded in the rendering pipeline, intercepting each G-buffer colour frame and producing a stylised version that is then passed through the Post-process stage. Figure~\ref{fig:our_results} demonstrates some final rendered frames for different reference style images and for different game scenes. Our method is capable of producing robust stylisations even for complicated scenes with difficult lighting. The final renders do not suffer from undesired artefacts or flickering effects, while the halo effect around the objects in the scene is significantly reduced. The following subsections further demonstrate temporally consistent stylisations of frames from various open-source games \cite{unity2022terrain,unity2022fontainebleau,polygonautic2020seed,unity2023book} and compare the results against state-of-the-art methods in image and video style transfer. Videos of our results and comparisons to state-of-the-art methods are included on the project's website. 

\subsection{Qualitative Results}

\begin{figure*}[htb]
    \begin{center}\includegraphics[width=\linewidth]{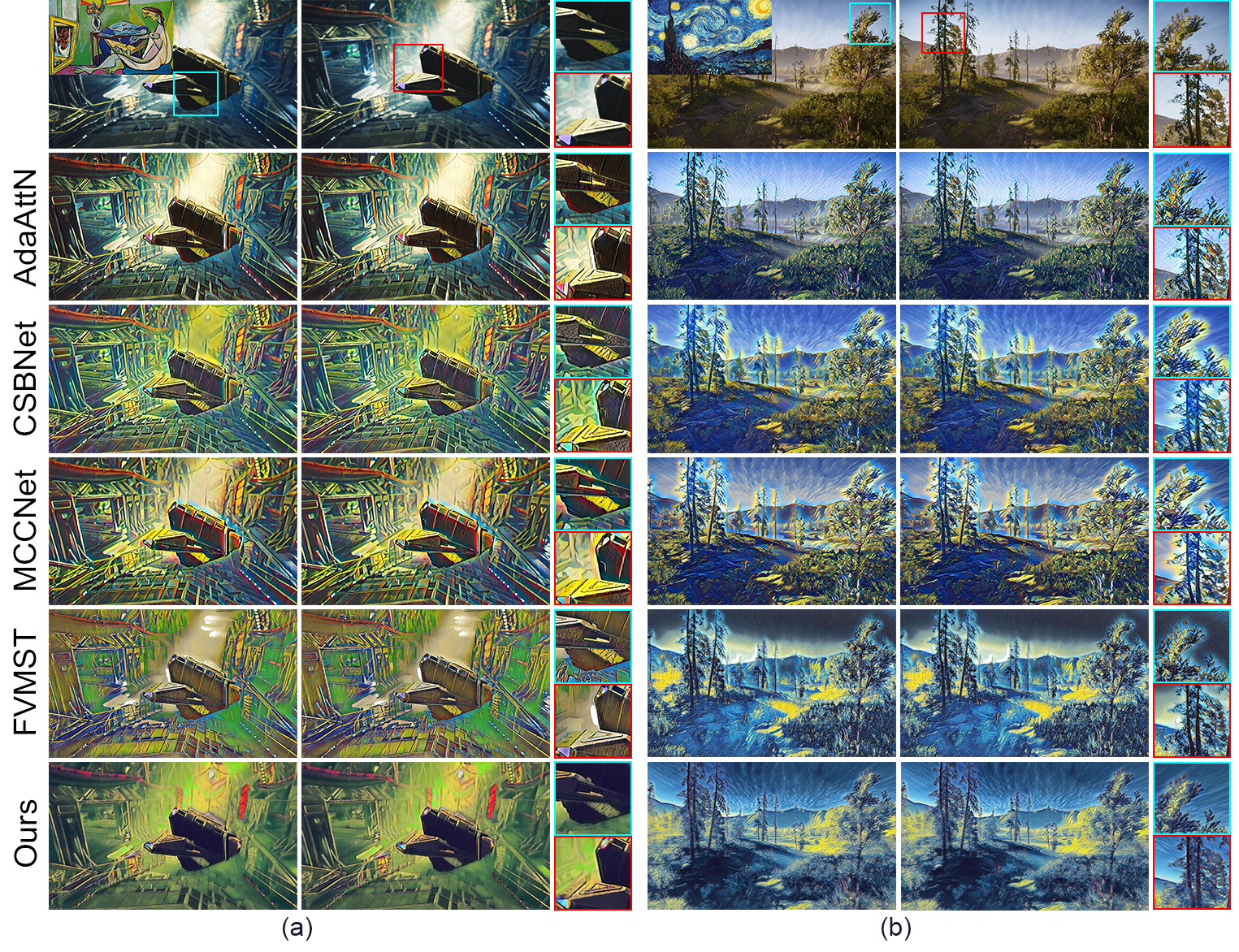}
    \end{center}
   \caption{Comparison against state-of-the-art approaches. Top row: original frames, with the style image top left; two adjacent frames from two different game scenes are shown. We provide zoomed-in cut-outs on the right of each two-frame sequence, for better comparisons. Our method produces robust stylisations that capture the style image more efficiently and preserve content and luminance information of the scene more effectively in comparison with the state-of-the-art approaches.}
\label{fig:results_comparison_2}
\end{figure*}

Qualitative comparisons against four state-of-the-art methods -- AdaAttN~\cite{liu2021adaattn}, CSBNet~\cite{lu2022universal}, MCCNet~\cite{deng2021arbitrary}, and FVMST~\cite{gao2020fast} -- are shown in Figure~\ref{fig:results_comparison_2}. This includes stylisations for two consecutive frames for two different game scenes and two different style images. 

Figure ~\ref{fig:results_comparison_2} shows that AdaAttN \cite{liu2021adaattn} preserves much of the content information, however, the stylisation effect is not very visible -- the yellow colour that is eminent in the style image is absent from the stylised frames (Figure~\ref{fig:results_comparison_2}(b)). The video NST methods, CSBNet~\cite{lu2022universal} and MCCNet~\cite{deng2021arbitrary}, reproduce the style image more faithfully than AdaAttN, but they create undesired artefacts such as the yellow halo around the trees that are visually distinguished from the background. FVMST \cite{gao2020fast} also captures the style quite well, but generates a white artefact encircling the mountain's background edge and it produces a sudden shift of the sky colour (from light, it turns to dark blue/black) that is not visible in the original frames. Our approach reproduces the style image faithfully and eliminates the undesired effects that are visible in the results of the state-of-the-art methods. The structure of the original frames is preserved comparably to the stylisations of AdaAttN but with higher stylisation intensity and better preservation of the luminance and lighting of the scene.

To demonstrate the effectiveness of our approach, we also include close-ups of consecutive frames of a game scene that includes a moving 3D object in a complex background. When looking at the zoomed-in cut-outs in Figure~\ref{fig:results_comparison_2}(a), the halo effect around the object's edge is more noticeable in the generated frames of AdaAttN~\cite{liu2021adaattn}, CSBNet~\cite{lu2022universal}, and MCCNet~\cite{deng2021arbitrary}, while the stylisations of FVMST~\cite{gao2020fast} create a disturbing white blob. In addition, the close-ups provide strong evidence of the capability of our approach to retaining the luminance in the scene and the game's post-effects. The prominent depth-of-field effect in the original frames is completely ignored when the stylisation is performed at the final render using state-of-the-art methods that enhance the details on the background. Our system makes the 3D object stand out and preserves the lighting and the game's overall look and feel.

\subsection{Quantitative Results}
\label{sec:quantitative}

For quantitative comparisons, frames are extracted 
from 4 different games and 12 different gameplays, including indoor and outdoor scenes, featuring moving objects and complicated lighting. This results in an evaluation dataset of 2100 frames (9 gameplays $\times$ 200 frames and 3 gameplays $\times$ 100 frames), and we evaluate utilising 10 different style images. The average results are reported in Table~\ref{tab:results_temp_style}. 

To quantitatively gauge the performance of our method in video stability and temporal coherence we utilise the warping error that is calculated as the difference between a warped next frame (using optic flow) and the original next frame. \textit{FlowNetS}~\cite{ilg2017flownet} is used to compute the optical flow of the original videos. In addition, we employ the LPIPS (Learned Perceptual Image Patch Similarity) metric \cite{zhang2018unreasonable} to measure the average perceptual distances between the adjacent frames in order to verify the smoothness of the stylised game sequences. The results show that our approach is superior to the state-of-the-art methods in generating temporally consistent in-game stylisations.

\begin{table}[htb]
\begin{center}
\begin{tabular}{|l|c|c|| c | c | c | c |}
    \hline
    Method & Warping Error~$\downarrow$ & LPIPS Error~$\downarrow$ & SSIM~$\uparrow$ & SIFID~$\downarrow$ & $\mathcal{L}_{c}$~$\downarrow$ & $\mathcal{L}_{s}$~$\downarrow$ \\
    \hline\hline
    AdaAttN~\cite{liu2021adaattn} & 1.6477 & 0.3217 & \textbf{0.7820} & 1.6115 &	\textbf{0.4945}	& 	1.0391				
 \\
    CSBNet~\cite{lu2022universal} & 1.7458 & 0.3908 & 0.6370 & 2.2468 & 0.8674 & 1.0053 \\
    MCCNet~\cite{deng2021arbitrary} & 1.6519 & 0.3547 & 0.6637 & 1.5555 &  0.8065 & 1.0042  \\
    FVMST~\cite{gao2020fast} & 1.8524 & 0.3215 & 0.5855 & 2.2529 & 0.7834 & 1.0077 \\
    \hline
    Ours (image) & 1.6764  & 0.3602 & 0.6740 & \textbf{1.2063} & 0.6532 & \textbf{0.9808} \\
    Ours (in-game) & \textbf{1.5798} & \textbf{0.2930} & 0.6057  & 1.8679 &  0.7830 & 1.0612  \\
    \hline
    \end{tabular}
\end{center}
\caption{Quantitative results. Warping Error and LPIPS error (both in the form $\times 10$) capture the smoothness of the generated video. SSIM and $\mathcal{L}_{c}$ relate to content preservation, and SIFID and $\mathcal{L}_{s}$ quantify the style performance. Results are given for our NST system injected in the game's rendering pipeline (game) and for the NST network applied as a post-effect (image). Bold values in Warping Error and LPIPS Error indicate our (in-game) approach is best at preserving temporal consistency.}
\label{tab:results_temp_style}
\end{table}

Perceptual metrics are employed to quantitatively assess the stylisation quality.
SSIM~\cite{wang2004image} and Content error ($\mathcal{L}_{c}$) \cite{gatys2016image} are used to evaluate the effectiveness of the methods in retaining content information; SIFID \cite{shaham2019singan} and Style error \cite{gatys2016image} are used to evaluate style performance. Our system manages to preserve content adequately. Whilst our algorithm's effectiveness in reproducing the style image is sufficient, some stylisation qualities are lost when the post-process effects are performed on top of the stylisations. In order to retain the intended post-effects applied to a game, certain aspects of the style likeness to the original image are traded off. Arguably, this compromise can be deemed desirable in a game setting and, as has been demonstrated, this trade-off leads to more consistent and temporally stable stylisations.

\subsection{Ablation Study}

\begin{figure*}[htb]
    \begin{center}
        \includegraphics[width=\linewidth]{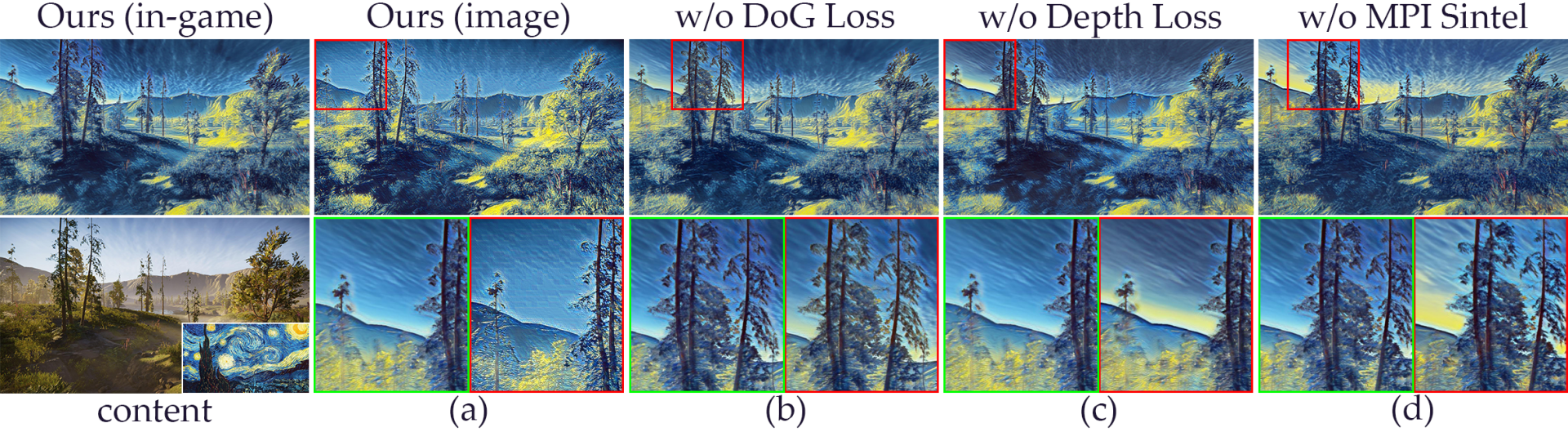}
    \end{center}
   \caption{Ablation study on the impact of the different components of our system. Each column shows a zoomed-in comparison between our method (in-game) trained with all components (green) and our stylisation network (a) applied as a post-effect, (b) trained without DoG Loss (c) trained without Depth Loss, and (d) trained without the MPI Sintel data (red).} 
\label{fig:ablations}
\end{figure*}

Figure~\ref{fig:ablations} demonstrates example results of our approach under different configurations. In-game stylisation has a significant impact on temporal coherence, in comparison to stylising each rendered frame as a post-effect (Section~\ref{sec:quantitative}). Here, we show that the latter also produces less appealing stylisations, with much of the content information being discarded. In addition, training without the DoG Loss results in more visible halos around the objects in the scene, whereas training with DoG Loss leads to generated frames with enhanced object stylisation and reduced boundary artefacts. The same applies to Depth Loss, as our method synthesises visibly improved results when depth is considered. The inclusion of the MPI Sintel dataset also has an impact on the performance -- the stylisation network trained only on the MS COCO dataset neglects the synthetic nature of the game and struggles to generate frames that retain the content adequately, producing undesired effects.

\subsection{Limitations}

To demonstrate the effectiveness of applying NST as part of the rendering pipeline of a computer game, we have trained a single-style-per-network model. Future work could experiment with arbitrary-style-per-model networks \cite{huang2017arbitrary,liu2021adaattn,deng2021arbitrary} which would provide the user with the option to upload and use their own reference style image. Another important consideration in applying NST in a game setting is running time. We reduced the number of residual layers and removed activation from the initial convolution layers to improve upon the inference time of the trained network which requires approximately 0.9 seconds to stylise an image of size 512 $\times$ 512. When injecting stylisation in the rendering pipeline the frame rate of a game running in Unity at Full HD resolution drops to $\sim$10fps. Utilising a more lightweight network architecture (arbitrary style transfer networks report better inference time, e.g., AdaIN \cite{huang2017arbitrary}: 0.065 seconds) could result in stylised game environments running at higher frame rates.

\section{Conclusion}
\label{sec:conclusions}

We have proposed a novel approach for injecting NST into a computer graphics rendering pipeline. Our NST framework is capable of producing coherent and temporally stable stylised frames in computer games. Our NST module intercepts frames from the colour buffer and synthesises artistic stylisations that are then written back to the camera colour buffer. Robust stylisations are achieved without interfering with the applied post-process effects. We demonstrate qualitative and quantitative results that reveal a promising new avenue for integrating NST within game development processes.

\section*{Acknowledgements}
This research was funded by the EPSRC.

\bibliographystyle{unsrt}  
\bibliography{references}

\end{document}